\pdfoutput=1

\documentclass[11pt, table]{article}

\usepackage[]{emnlp2021}

\usepackage{times}
\usepackage{latexsym}

\usepackage[T1]{fontenc}

\usepackage[utf8]{inputenc}
\usepackage{graphicx}
\usepackage{multirow}
\usepackage{verbatim}
\usepackage{lipsum}
\usepackage{amssymb}
\usepackage{amsmath}
\usepackage{booktabs}
\usepackage{url} 
\usepackage{enumitem}
\usepackage{mathtools}
\usepackage{subcaption}
\usepackage{fontawesome} 
\usepackage{contour}
\usepackage{textcomp}

\usepackage{microtype}

\title{Probing Commonsense Explanation in Dialogue Response Generation}

\author{
Pei Zhou \quad Pegah Jandaghi \quad Hyundong Cho \quad Bill Yuchen Lin  \\ \textbf{Jay Pujara \quad Xiang Ren}\\
Department of Computer Science and Information Sciences Institute\\
University of Southern California\\
\small{\texttt{\{peiz,yuchen.lin,jpujara,xiangren\}@usc.edu}, \texttt{\{jandaghi,jcho\}@isi.edu}}
}

\begin{document}
\maketitle
\begin{abstract}



Humans use commonsense reasoning (CSR) implicitly to produce natural and coherent responses in conversations. 
Aiming to close the gap between current response generation (RG) models and human communication abilities, we want to understand \emph{why} RG models respond as they do by probing RG model's understanding of commonsense reasoning that elicits proper responses.
We formalize the problem by framing commonsense as a latent variable in the RG task and using explanations for responses as textual form of commonsense. We collect 6k annotated \emph{explanations} justifying responses from four dialogue datasets and ask humans to verify them and propose two probing settings to evaluate RG models' CSR capabilities. Probing results show that models fail to capture the logical relations between commonsense explanations and responses and fine-tuning on in-domain data and increasing model sizes do not lead to understanding of CSR for RG. 
We hope our study motivates more research in making RG models emulate the human reasoning process in pursuit of smooth human-AI communication~\footnote{Our code and data are on our project page:~\url{https://sites.google.com/usc.edu/cedar}.}.
\end{abstract}

\section{Introduction}\label{intro}






Response generation (RG) systems, which have the basic goal of mimicking human conversation, have as of yet an unmeasured ability to understand communicative intents. In general, standard neural language models build correlative models of linguistic stimuli rather than deep understanding of human-level meaning~\cite{bender-koller-2020-climbing}. As such, there is reason to suspect that, while RG systems today have impressive performance on common metrics~\cite{zhang2020dialogpt,roller2020recipes}, they achieve this performance without truly understanding human communication. 
Commonsense reasoning (CSR), defined as ``\emph{the basic level of practical knowledge and reasoning concerning everyday situations and events that are commonly shared among most people}''~\cite{sap2020commonsense}, is critical in human communication. Specifically, CSR helps establish a common ground consisting of ``\emph{mutual knowledge}'' between participants, which is key to smooth communication~\cite{clark1989contributing, clark1991grounding}.

\begin{figure}[t]
	\centering
	\includegraphics[width=0.8\columnwidth]{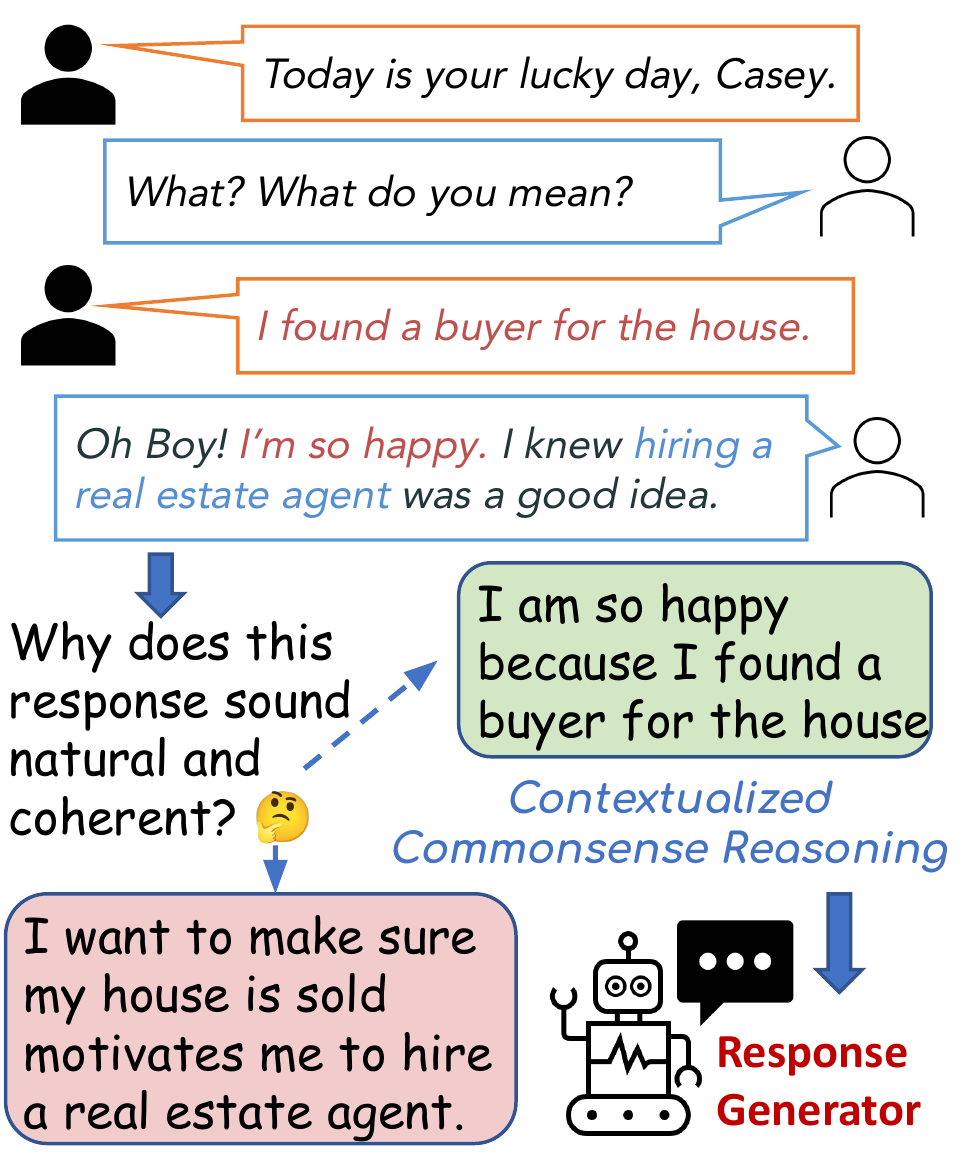}
	\caption{
	{\textbf{A motivating example for our study.} We want to know whether RG models understand the implicit common sense that justifies dialogue responses.}}
	\label{fig:motivating}
\end{figure}



For example, consider a conversation between two friends shown in Figure~\ref{fig:motivating}. The reason the person on the right (responder) is happy is not indicated explicitly, but it is common sense that finding a buyer for the house (that the responder is likely aiming to sell) makes one happy, which explains the response ``\emph{I'm so happy}''. 
Motivated by how humans communicate, we ask a main research question: \emph{do RG models understand the implicit CSR that explains why a response makes sense?} This will help us analyze whether the RG models that seem to produce human-like responses really understand the reasoning process that justifies the response, which is important to build a reliable and robust dialogue system. Furthermore, understanding implicit common sense behind RG can also help make models generate more natural and coherent responses.
To answer this important research question, we present our initial findings from \emph{annotating} commonsense explanations in dialogues and \emph{evaluating RG models} for commonsense reasoning capabilities.

We first present a probing setup for evaluating common sense in RG, called \emph{CEDAR}: \textbf{C}ommon s\textbf{E}nse in \textbf{D}i\textbf{A}logue \textbf{R}esponse generation.
We start with formalizing CSR in RG by considering common sense as a latent variable that helps explain the observed variable ``\emph{response}'' in the RG process -- similar to how humans use common sense in communication~\cite{hilton1990conversational}.
To instantiate implicit common sense for probing, we use textual explanations of the response as the common sense embedded in the dialogue context. 
To understand whether RG models can comprehend implicit common sense, we \emph{corrupt} explanations to break the logical coherence and compare model behaviors between a valid explanation and a corrupted one.

To operationalize the probing, we collect the first annotations on commonsense explanations that justify dialogue responses. 
Each annotation is a dialogue-specific explanation that explicitly describes what might cause the response in one of the five dimensions: event, emotion, location, possession, and attribute, inspired by human cognitive psychology~\cite{kintsch1978toward}.
We find through pilot studies that directly asking people to annotate result in explanations with high variation and subjectivity, to account for this, we first generate candidate explanations by adopting a large text-to-text language model trained on a story explanation dataset, namely GLUCOSE~\cite{mostafazadeh2020glucose}, under the dialogue setting. Next, we conduct a carefully designed two-stage human verification process with a qualification test and the main annotation task. We present our findings from verifying 6k generated explanations on 1,200 dialogues sampled from four public dialogue datasets. 

Using the annotated explanations, we probe
state-of-the-art (SOTA) RG models for two CSR-related abilities: (i) the ability to understand whether the commonsense explanation can justify a response, and (ii) the ability to attribute logically-coherent explanations for dialogue responses. These are inspired by what showcases human understanding of common sense in conversational communication. Our probing setup contrasts valid explanations with corrupted version. Corruptions are generated via two methods: \textit{logical corruptions} that disrupt logical coherence, and \textit{complete corruption} where we disrupt the grammatical naturalness of the sentence.

We find that the models fail to understand common sense that elicits proper responses according to performance on our probing settings and some models even do not distinguish gibberish sentences. Fine-tuning on in-domain dialogues and verified explanations do not help with understanding. We also find interesting cases that show potential statistical biases in RG models. We hope our annotated explanations and probing findings encourage more studies on making RG models communicates with deep understanding of human reasoning process.



\section{Task Formulation and Challenges}\label{task_challenge}
This section first introduces how we incorporate common sense as a latent variable in the RG setting. Then we specify two challenges that arise in order to examine whether RG models can comprehend common sense to arrive at responses similarly as humans do. Lastly, we present our solutions to the challenges by instantiating common sense as textual explanation and proposing two probing settings to evaluate if models reason about common sense when generating responses.





\subsection{Common Sense in Response Generation}

\paragraph{Preliminaries} 
We consider the classic dialogue response generation (RG) setup~\cite{weizenbaum1966eliza, ritter-etal-2011-data, sordoni2015neural}: given a dialogue \emph{history} $H$, generate an appropriate \emph{response} $R$.
Most state-of-the-art (SOTA) neural RG models generate a response given a dialogue history as a \emph{conditional language modeling} problem. Specifically, given a \emph{history ($H$)} consisting of a sequence of dialogue turns from the dialogue history $x_1, x_2, ..., x_n$ (each containing a sequence of tokens) and a \emph{response ($R$)}  sentence $y$ comprised of a sequence of tokens $y_1, y_2, ..., y_m$, RG models aim to learn the conditional probability distribution by training on human dialogues:
\begin{equation}
    P_{\theta}(R|H)=\prod_{i=1}^{m}P_{\theta}(y_i|y_{<i}, x_1,...,x_n).
\end{equation}

\begin{figure}[t]
	\centering
	\includegraphics[width=0.8\columnwidth]{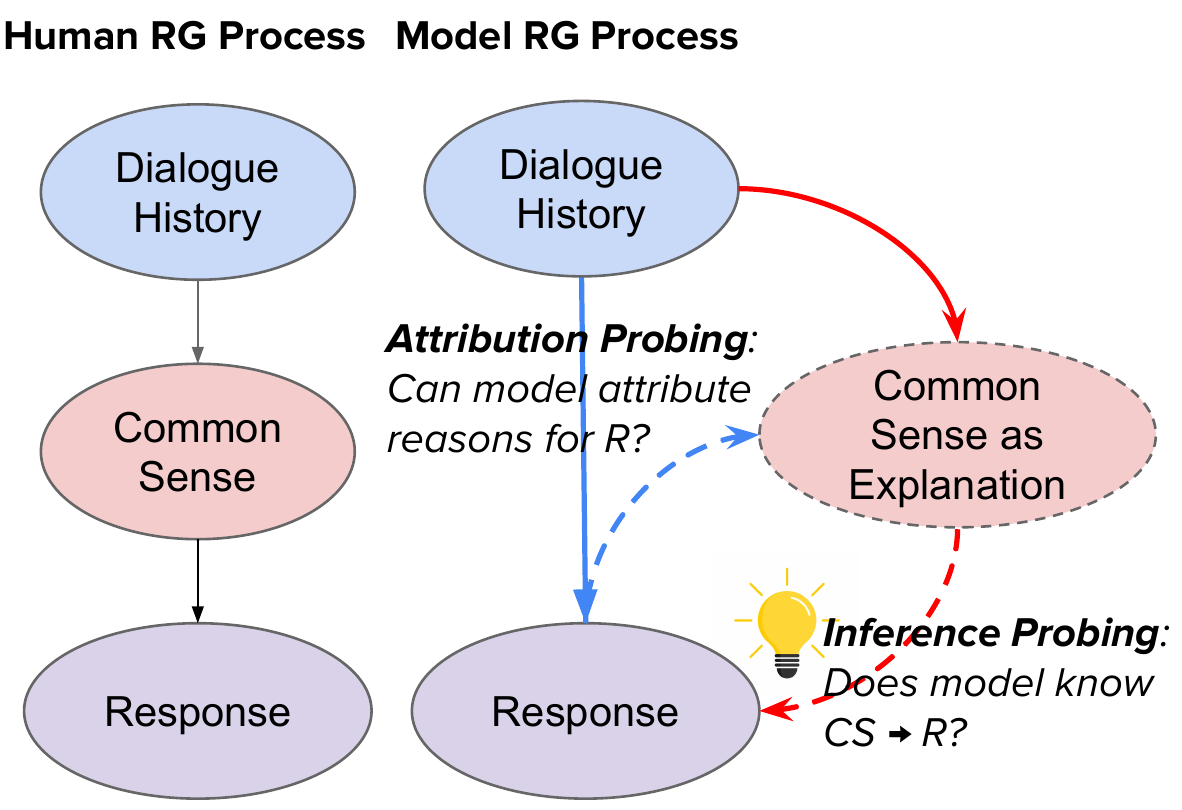}
	\caption{
	\textbf{Probing setting} illustrations. We draw inspirations from human reasoning process during communication and probe RG models' understanding of implicit common sense in RG in two ways (red and blue dotted lines).
	}
	\label{fig:probing}
\end{figure}

\paragraph{Common Sense as a Latent Variable}
As illustrated in Figures~\ref{fig:motivating} and \ref{fig:probing}, when humans respond in a conversation, we use common sense implicitly to establish \emph{common ground}~\cite{grice1975logic, clark1991grounding}, reach mutual understanding, and help produce natural responses for smooth communication. 
We consider common sense to be \textit{latent} because it is infrequently stated due to the cooperative principle that states that participants should ``\emph{not make your contribution more informative than is required}''~\cite{grice1975logic}. However, the reasoning it enables is an integral part establishing common ground and critical for communication.
To formalize this process, we consider
\emph{common sense} ($CS$) as an important \textit{latent variable} in the modeling of a dialogue response when given the history -- \textit{i.e}, $P(R|H,CS)$. Other latent factors such as the environment in which the conversation happens and background information of the participants can also influence the dialogue, but here we focus on common sense.

\subsection{Probing Setup}\label{sec:task_setup}

Current RG models generate responses in an end-to-end manner with only input from dialogue history (\textit{i.e.}, $H$), making it non-trivial to examine if they understand the implicit common sense behind RG process (also see Figure~\ref{fig:probing} for an illustration). 
We instantiate implicit common sense in dialogues and then design probes to evaluate models' grasp of common sense. This leads to two key challenges: 1) \emph{how to instantiate abstract and implicit common sense $CS$ in dialogues?} and 2) \emph{how to probe RG models' understanding of common sense in dialogue response generation?}

\paragraph{Instantiate Common Sense Using Explanations}
We use \emph{natural language explanations justifying why a response makes common sense} as a proxy to instantiate common sense in RG.
Traditional studies have tied common sense and the ability to provide explanations for events and actions closely~\cite{hansen1980commonsense,hilton1986knowledge}, which also holds true in a conversational setting~\cite{hilton1990conversational}.
Specifically, as shown in Figure~\ref{fig:motivating}, ``\emph{I want to make sure my house is sold}'' is a potential explanation about what leads to ``\emph{hiring a real estate agent}'' in the response and this explanation requires understanding the commonsense relation that a real estate agent helps sell a house and the desire to sell a house motivates a person to hire an agent. 
Formally, we concretize the abstract latent variable common sense $CS$ in textual form as an explanation $E$ explaining what might cause the response $R$ given the history $H$.
We introduce our process of collecting such explanations for RG in Section~\ref{annotation}.

\paragraph{Probe Models' Understanding in Two Settings}
We then draw inspiration from human reasoning process behind dialogue response generation to design two probing tasks.
First, humans use common sense implicitly to produce natural and coherent responses in conversations~\cite{clark1989contributing}. Common sense helps humans determine what responses make sense in certain context. 
We want to see if providing common sense in the form of explanation also helps RG models arrive at coherent and natural responses more easily. 
Second, humans can perform \emph{causal attribution} on an event or an action by finding reasons that might cause it~\cite{hilton1990conversational}. If the person producing the response is asked about why they are feeling happy, they can easily respond with reasons about their reasoning process. We are interested in examining can RG models also generate responses to justify a previous response when asked. 

We probe RG models in a \emph{contrastive} manner, by comparing model behaviors with a valid explanation $E$ to the response and a \emph{corrupted} $E'$ that breaks logical coherence. We introduce the two settings in more detail as follows.

\paragraph{Inference Probing} 
Here we directly measure if $P(R|E,H)>P(R|E',H)$ for RG models, \textit{i.e.}, can models assign a higher probability to the response when provided with valid common sense in the form of explanations compared to logically-incoherent explanations? 
Since existing RG models are not trained to take explanations as additional input, the probing results may be confounded by the model's unfamiliarity with the probing setting. To account for this issue, we 1) probe on a knowledge-grounded RG model that is used to taking in additional knowledge sentences as input and 2) fine-tune RG models on a proportion of our collected explanation and compare the effects. 
We discuss results and issues about probing models fine-tuned on explanations in Section~\ref{Sec:analysis_results}.

If the model assigns a similar or lower probability to the response given a valid explanation compared to a logically-incoherent explanation, it indicates that the reason why this response makes sense is not clear for models.

\paragraph{Attribution Probing} 
Here we examine if $P(E|H,R)>P(E'|H,R)$, \textit{i.e.}, can RG models perform \emph{causal attribution} as humans by assigning a higher probability to a valid explanation of the response (that makes sense) compared to a corrupted explanation, given the dialogue history and the response? 
To address the unfamiliarity of models, we make the probing setting close to real dialogues by continuing the conversation (consisting of $H$ and $R$) with ``\textit{why}'' to prompt the models to generate an explanation. 
We also conduct fine-tuning on a proportion of our collected explanations similarly to the first setting discussed in Section~\ref{Sec:analysis_results}.

If the model prefers the attribution of the response that is incoherent with the response by giving it a higher probability, it indicates that the model fails to generate valid reasons for responses, which requires understanding the implicit common sense behind dialogues.

\section{Generating Commonsense Explanations for Dialogue Responses}\label{annotation}

To get explanation annotations for dialogue responses, we first automatically generate commonsense causal explanations and then manually verify via crowdsourcing.
We use a text-to-text model trained on commonsense story explanation dataset GLUCOSE~\cite{mostafazadeh2020glucose} as the generator and conduct 2-stage human verification on generated explanations. We first introduce the model we use, the adaptation of the model on dialogue data, and our verification process to ensure the quality of generated explanations.


\subsection{Generating Commonsense Explanations}


GLUCOSE is a large-scale dataset of implicit commonsense causal explanations grounded in a story context~\cite{mostafazadeh2020glucose}. Given a short story and a sentence X in the story, GLUCOSE contains human annotations of five dimensions of causal explanation related to X (an event/emotion/location/possession/attribute leads to X), each in a semi-structured form ``antecedent \emph{connective} consequent.'' 
Using the collected explanations, the authors train state-of-the-art neural models and find that the trained models are able to produce commonsense inferences on unseen stories.  More details about how models are trained on GLUCOSE are included in Appendix~\ref{glucose}.


\begin{figure}[t]
	\includegraphics[width=\columnwidth]{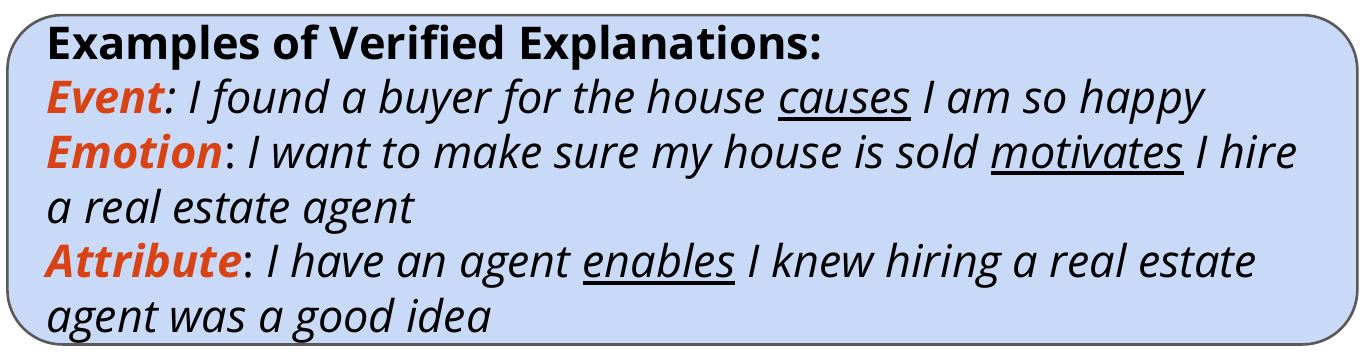}
	\caption{
	{Examples of human-verified commonsense explanations for the dialogue shown in Figure~\ref{fig:motivating}}}
	\label{fig:verified}
\end{figure}

We consider using a model trained on GLUCOSE to automatically generate commonsense explanations in dialogues for several reasons. First, it generates \emph{contextual commonsense explanations} that provides causal knowledge about what justifies a sentence. Second, it provides fine-grained causal explanations along different dimensions. Last but not least, we have conducted multiple rounds of pilot studies to directly ask workers to \emph{write out} commonsense explanations for a response, but the subjectivity of this open-ended task led to large variations in quality. 
Instead we ask workers to \emph{verify} explanations generated from a model.

We sample 1,200 dialogues from 4 dialogue datasets (300 from each): DailyDialog~\cite{li2017dailydialog}, EmpatheticDialogues~\cite{rashkin2019towards}, MuTual~\cite{cui2020mutual}, and SocialIQA-prompted dialogues~\cite{zhou-etal-2021-commonsense}. We generate 6k commonsense causal explanations (5 dimensions for each dialogue), using the last turn as the response and the previous turns as dialogue history (after filtering short turns). We follow ~\citet{zhou-etal-2021-commonsense}'s approach to select dialogues that contain at least a one-hop triple from ConceptNet~\cite{liu2004conceptnet}. We use the same hyperparameters and weights from the best-performing 770M T5 model from~\citet{mostafazadeh2020glucose}.

\subsection{Verification}
To ensure the quality of generated explanations, we carefully design a two-stage human \emph{verification} process with a qualification test and the main task. 
Workers must first pass a \emph{qualification test} (QT) that tests their understanding of the CS criteria necessary for  our main annotation tasks (more details in Appendix~\ref{sec:veri_detail}).
We consider three criteria, requiring generated explanations to pass all three to be considered a valid commonsense explanation for a response.
We ask \emph{three} workers on Amazon Mechanical Turk (MTurk) to annotate the \emph{three} criteria for each explanation. 

\paragraph{Criteria}
1). \textbf{Relevant}. A good causal explanation has to focus on explaining what could cause the \emph{response} in the dialogue context~\cite{hilton1990conversational}. 
An example of an irrelevant explanation for the example shown in Figure~\ref{fig:motivating} is ``\emph{I possess a house enables I live in a house}'' since ``\emph{living in a house}'' is not what the response is about, so it doesn't help explain the response.
2). \textbf{Non-trivial}. We observe that sometimes the model simply duplicates a previous dialogue turn as the cause, which trivially associates history and response. We are interested in implicit and specific commonsense so we filter out explanations that parrot a previous turn. For example, ``\emph{I found a buyer for the house motivates Oh Boy! I’m so happy. I knew hiring a real estate agent was a good idea.}''
3). \textbf{Plausible}. We ask humans to verify if the generated explanation plausibly identifies a likely cause for the response. 
An example of an implausible explanation is ``\emph{I am in a house enables I am so happy}'' since ``\emph{I am in a house}'' is not the direct cause why the person producing the response is feeling so happy, ``\emph{found a buyer for the house}'' is. This is the hardest criterion for humans to decide due to its subjectivity nature.

\paragraph{Results}
We present results of our verification of 6k explanations from three in-house annotators. 
To filter ambiguous explanations and be strict about the quality of verified explanations, we only consider explanations valid if \emph{all} three annotators have agreed that they satisfy \emph{all} three criteria, i.e., 100\% agreement for all verified explanations.
For the annotated explanation, passing rates (agreed by 3 workers) for criterion (relevant, non-trivial, plausible) are (55\%, 73\%, 37\%) -- yielding an overall passing rate of 26\% (1,560 explanations).
Passing rates for the five dimensions (event, emotion, location, possession, attribute) are (31\%, 33\%, 13\%, 24\%, and 29\%), with more details in Appendix~\ref{sec:veri_detail}.
Figure~\ref{fig:verified} presents examples for different dimension, full data is included in the supplementary material.




\section{Probing Setup}\label{probing}

\begin{table*}[tb]
\centering
\scalebox{0.7}{
\begin{tabular}{ccccccccc}
\hline
\multicolumn{1}{c|}{}                         & \multicolumn{4}{c|}{\textbf{Logical Corruption Average {[}Accuracy/$\Delta$ NLL{]}}}                                                                                                                   & \multicolumn{4}{c}{\textbf{Complete Corruption Average {[}Accuracy/$\Delta$ NLL{]}}}                                                                                              \\ \cline{2-9} 
\multicolumn{1}{c|}{\multirow{-2}{*}{Models}} & DD                                         & ED                                         & MuTual                                     & \multicolumn{1}{c|}{SocialIQA}                                  & DD                                         & ED                                         & MuTual                                     & SocialIQA                                  \\ \hline
\multicolumn{9}{c}{\cellcolor[HTML]{C0C0C0}\textit{\textbf{Inference Probing}}}                                                                                                                                                                                                                                                                                                                                                            \\ \hline
\multicolumn{1}{c|}{DialoGPT (l2r)}           & \cellcolor[HTML]{CFE2F3}0.57/-0.01         & \cellcolor[HTML]{CFE2F3}\textbf{0.60/0.03} & \cellcolor[HTML]{CFE2F3}\textbf{0.62/0.03} & \multicolumn{1}{c|}{\cellcolor[HTML]{CFE2F3}\textbf{0.64/0.03}} & \cellcolor[HTML]{FCE5CD}0.71/0.15          & \cellcolor[HTML]{FCE5CD}0.77/0.25          & \cellcolor[HTML]{FCE5CD}\textbf{0.79/0.22} & \cellcolor[HTML]{FCE5CD}\textbf{0.87/0.40} \\
\multicolumn{1}{c|}{TopicalChat-GPT2 (l2r)}   & \cellcolor[HTML]{CFE2F3}0.49/-0.00         & \cellcolor[HTML]{CFE2F3}0.50/-0.00         & \cellcolor[HTML]{CFE2F3}0.49/-0.00         & \multicolumn{1}{c|}{\cellcolor[HTML]{CFE2F3}0.50/-0.00}         & \cellcolor[HTML]{FCE5CD}\textbf{0.76/0.23} & \cellcolor[HTML]{FCE5CD}\textbf{0.79/0.24} & \cellcolor[HTML]{FCE5CD}0.78/0.24          & \cellcolor[HTML]{FCE5CD}0.81/0.27          \\
\multicolumn{1}{c|}{BlenderBot (s2s)}         & \cellcolor[HTML]{CFE2F3}0.46/0.00          & \cellcolor[HTML]{CFE2F3}0.55/0.02          & \cellcolor[HTML]{CFE2F3}0.51/0.02          & \multicolumn{1}{c|}{\cellcolor[HTML]{CFE2F3}0.50/0.01}          & \cellcolor[HTML]{FCE5CD}0.45/-0.02         & \cellcolor[HTML]{FCE5CD}0.43/-0.05         & \cellcolor[HTML]{FCE5CD}0.49/-0.03         & \cellcolor[HTML]{FCE5CD}0.41/-0.03         \\
\multicolumn{1}{c|}{BART-base (s2s)}          & \cellcolor[HTML]{CFE2F3}\textbf{0.53/0.07} & \cellcolor[HTML]{CFE2F3}\textbf{0.60/0.19} & \cellcolor[HTML]{CFE2F3} 0.57/0.07 & \multicolumn{1}{c|}{\cellcolor[HTML]{CFE2F3}0.54/0.09} & \cellcolor[HTML]{FCE5CD}0.36/-0.38         & \cellcolor[HTML]{FCE5CD}0.41/-0.23         & \cellcolor[HTML]{FCE5CD}0.43/-0.27         & \cellcolor[HTML]{FCE5CD}0.43/-0.21         \\
\multicolumn{1}{c|}{BART-large (s2s)}         & \cellcolor[HTML]{CFE2F3}0.51/-0.03         & \cellcolor[HTML]{CFE2F3}0.52/-0.01         & \cellcolor[HTML]{CFE2F3}0.48/-0.06         & \multicolumn{1}{c|}{\cellcolor[HTML]{CFE2F3}0.52/0.00}          & \cellcolor[HTML]{FCE5CD}0.49/-0.05         & \cellcolor[HTML]{FCE5CD}0.55/0.06          & \cellcolor[HTML]{FCE5CD}0.52/0.01          & \cellcolor[HTML]{FCE5CD}0.57/0.11          \\ \hline
\multicolumn{1}{c|}{DialoGPT-ft (l2r)}        & \cellcolor[HTML]{CFE2F3}0.50/-0.05         & \cellcolor[HTML]{CFE2F3}0.39/-0.54         & \cellcolor[HTML]{CFE2F3}0.44/-0.33         & \multicolumn{1}{c|}{\cellcolor[HTML]{CFE2F3}0.43/-0.25}         & \cellcolor[HTML]{FCE5CD}0.63/0.11          & \cellcolor[HTML]{FCE5CD}0.76/0.24          & \cellcolor[HTML]{FCE5CD}0.66/0.15          & \cellcolor[HTML]{FCE5CD}0.78/0.31          \\
\multicolumn{1}{c|}{BART-base-ft (s2s)}       & \cellcolor[HTML]{CFE2F3}0.59/0.02 & \cellcolor[HTML]{CFE2F3}0.58/0.01          & \cellcolor[HTML]{CFE2F3}0.58/0.02          & \multicolumn{1}{c|}{\cellcolor[HTML]{CFE2F3}0.60/0.03}          & \cellcolor[HTML]{FCE5CD}0.57/0.04          & \cellcolor[HTML]{FCE5CD}0.72/0.07          & \cellcolor[HTML]{FCE5CD}0.59/0.04          & \cellcolor[HTML]{FCE5CD}0.70/0.09          \\
\multicolumn{1}{c|}{BART-large-ft (s2s)}      & \cellcolor[HTML]{CFE2F3}0.57/0.02          & \cellcolor[HTML]{CFE2F3}0.44/-0.01         & \cellcolor[HTML]{CFE2F3}0.53/0.01          & \multicolumn{1}{c|}{\cellcolor[HTML]{CFE2F3}0.48/0.00}          & \cellcolor[HTML]{FCE5CD}0.35/-0.06         & \cellcolor[HTML]{FCE5CD}0.54/0.02          & \cellcolor[HTML]{FCE5CD}0.37/-0.04         & \cellcolor[HTML]{FCE5CD}0.48/-0.00         \\ \hline
\multicolumn{1}{c|}{Human}                    & \cellcolor[HTML]{CFE2F3}1.0                & \cellcolor[HTML]{CFE2F3}1.0                & \cellcolor[HTML]{CFE2F3}0.9                & \multicolumn{1}{c|}{\cellcolor[HTML]{CFE2F3}1.0}                & \cellcolor[HTML]{FCE5CD}1.0                & \cellcolor[HTML]{FCE5CD}1.0                & \cellcolor[HTML]{FCE5CD}1.0                & \cellcolor[HTML]{FCE5CD}1.0                \\ \hline
\end{tabular}
}
\caption{\textbf{Inference probing results} for different response generation models on 4 dialogues datasets against two categories of corruptions. Accuracy is the bianry accuracy of giving a lower loss to the valid explanation than the corrupted one and $\Delta$ NLL is the average difference of per-token NLL between the loss of a corrupted inference and a valid inference (the more positive the better).}
\label{tab:inf_results}
\end{table*}

\begin{figure}[t]
	\includegraphics[width=\columnwidth]{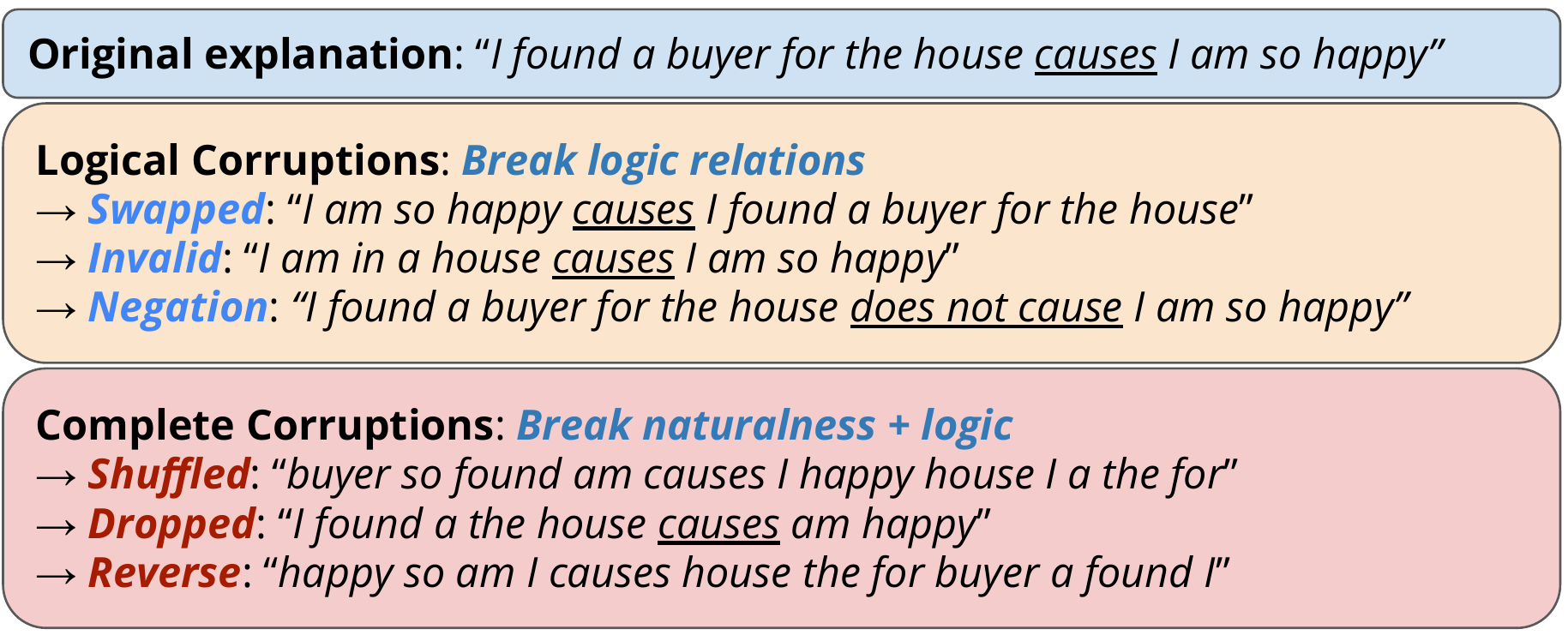}
	\caption{
	{Examples of different \textbf{corruption types} to a commonsense causal explanation.}}
	\label{fig:corruptions}
\end{figure}

We probe RG models' capability of understanding and using the explanation $E$ in a \emph{contrastive} manner (Sec.~\ref{sec:task_setup}).
This section first introduces the corruption types under two categories, then we introduce evaluation metrics, and finally we discuss several SOTA RG models with different neural architectures that we probe.


\subsection{Corruption Types}



We use verified explanations generated from GLUCOSE T5 model as valid explanations and define two categories of \emph{corruptions} to corrupt the explanation to be logically-invalid and/or grammatically unnatural. We consider three \emph{logical corruptions} that invalidate the logical connection between explanation and response, as well as three \emph{complete corruptions} that break both logical coherence and naturalness of the sentence. Examples covering showing corruption types of a valid explanation are shown in Figure~\ref{fig:corruptions}, for which ``\emph{I found a buyer for the house}'' is the antecedent, ``\emph{causes}'' is the connective, and ``\emph{I am so happy}'' is the consequent.

\paragraph{Logical Corruptions} We consider three ways to invalidate the logic of the explanation: 1) \emph{Swapped} that swaps the antecedent and consequent of the explanation, 2) \emph{Negation} that negates the connective word of the explanation, 3) \emph{Incorrect} that uses an explanation from the same dialogue history-response instance that is rated as incorrect (if any) during the verification.

\paragraph{Complete Corruptions} 
Inspired by \citet{sankar2019neural} who design perturbations to apply on dialogue history and analyze sensitivity of RG models by measuring the perplexity of the response, we consider three operations that completely break the naturalness of the explanation: 1) \emph{Shuffle} that randomly shuffles the words of the explanation, 2) \emph{Dropped} that drops 30\% of the words uniformly, 3) \emph{Reversed} reverses the ordering of all the words in the explanation.

\paragraph{Evaluation Protocol and Metrics}
We use two metrics to measure RG models' capability to distinguish valid commonsense causal explanations from invalid explanations. 
The standard way of modeling $P_{\theta}$ in Equation (1) in generative models is using Maximum Likelihood Estimation (MLE) approach and minimize the conditional negative log-likelihood loss (NLL), i.e., $L(P_{\theta},R,H) = - \sum_{i=1}^{m}\log P_\theta(y_i|y_{<i},x_1,...,x_n).$

Since NLL is a direct measure of the probability distribution learned by the models, we use the same NLL measure for probing RG models' behavior. 
To measure performance of RG models, we directly compare the average per-token NLL when given a valid explanation and when given an invalid explanation to the response.

We first consider binary accuracy of giving a lower loss (higher probability) to the valid explanation than the corrupted one. A random-guessing baseline for the accuracy is 0.5. To further measure how \emph{confident} the model is in determining the validity of commonsense explanations, we also compute the average difference $\Delta NLL$ by subtracting the loss of the valid inference from the invalid inference loss. The closer to zero the difference is, the less confident the model is. 

\subsection{Response Generation Models}
We experiment with multiple models from two neural architectures: GPT-2-based ~\cite{radford2019language} unidirectional transformer language model and Seq2Seq-based transformer~\cite{vaswani2017attention} models. For GPT-2-based models, we use \emph{DialoGPT} that is trained on 147M multi-turn conversation-like exchanges extracted from Reddit~\cite{zhang2020dialogpt} and GPT-2 trained on \emph{TopicalChat}~\cite{gopalakrishnan2019topical} as the knowledge-grounded RG model. For seq2seq models, we use BlenderBot~\cite{roller2020recipes} and BART~\cite{lewis2019bart}. More details about these RG models are included in Appendix~\ref{model_detail}.

\section{Probing Results and Analysis} \label{result}
							

\begin{table*}[tb]
\centering
\scalebox{0.75}{
\begin{tabular}{ccccccccc}
\hline
\multicolumn{1}{c|}{}                         & \multicolumn{4}{c|}{\textbf{Logical Corruption Average {[}Accuracy/$\Delta$ NLL{]}}}                                                                                                                   & \multicolumn{4}{c}{\textbf{Complete Corruption Average {[}Accuracy/$\Delta$ NLL{]}}}                                                                                              \\ \cline{2-9} 
\multicolumn{1}{c|}{\multirow{-2}{*}{Models}} & DD                                         & ED                                         & MuTual                                     & \multicolumn{1}{c|}{SocialIQA}                                  & DD                                         & ED                                         & MuTual                                     & SocialIQA                                  \\ \hline
\multicolumn{9}{c}{\cellcolor[HTML]{C0C0C0}\textit{\textbf{Attribution Probing}}}                                                                                                                                                                                                                                                                                                                                                          \\ \hline
\multicolumn{1}{c|}{DialoGPT (l2r)}           & \cellcolor[HTML]{CFE2F3}0.46/-0.07         & \cellcolor[HTML]{CFE2F3}0.47/-0.04         & \cellcolor[HTML]{CFE2F3}0.48/0.03          & \multicolumn{1}{c|}{\cellcolor[HTML]{CFE2F3}0.49/0.00}          & \cellcolor[HTML]{FCE5CD}0.91/1.60          & \cellcolor[HTML]{FCE5CD}0.93/2.32          & \cellcolor[HTML]{FCE5CD}0.92/1.90          & \cellcolor[HTML]{FCE5CD}0.93/2.36          \\
\multicolumn{1}{c|}{TopicalChat-GPT2 (l2r)}   & \cellcolor[HTML]{CFE2F3}0.57/0.05          & \cellcolor[HTML]{CFE2F3}0.55/0.10          & \cellcolor[HTML]{CFE2F3}0.57/0.10 & \multicolumn{1}{c|}{\cellcolor[HTML]{CFE2F3}0.55/0.09}          & \cellcolor[HTML]{FCE5CD}\textbf{0.97/2.75} & \cellcolor[HTML]{FCE5CD}\textbf{0.97/3.08} & \cellcolor[HTML]{FCE5CD}\textbf{0.96/2.93} & \cellcolor[HTML]{FCE5CD}\textbf{0.96/2.93} \\
\multicolumn{1}{c|}{BlenderBot (s2s)}         & \cellcolor[HTML]{CFE2F3}\textbf{0.60/0.04} & \cellcolor[HTML]{CFE2F3}\textbf{0.59/0.05} & \cellcolor[HTML]{CFE2F3}\textbf{0.60/0.05} & \multicolumn{1}{c|}{\cellcolor[HTML]{CFE2F3}\textbf{0.58/0.06}} & \cellcolor[HTML]{FCE5CD}0.83/0.45          & \cellcolor[HTML]{FCE5CD}0.87/0.72          & \cellcolor[HTML]{FCE5CD}0.86/0.58          & \cellcolor[HTML]{FCE5CD}0.84/0.55          \\
\multicolumn{1}{c|}{BART-base (s2s)}       & \cellcolor[HTML]{CFE2F3}0.39/-0.19         & \cellcolor[HTML]{CFE2F3}0.41/-0.14         & \cellcolor[HTML]{CFE2F3}0.44/-0.10         & \multicolumn{1}{c|}{\cellcolor[HTML]{CFE2F3}0.42/-0.13}         & \cellcolor[HTML]{FCE5CD}0.52/0.08          & \cellcolor[HTML]{FCE5CD}0.50/0.01          & \cellcolor[HTML]{FCE5CD}0.52/0.14          & \cellcolor[HTML]{FCE5CD}0.51/0.10          \\
\multicolumn{1}{c|}{BART-large (s2s)}         & \cellcolor[HTML]{CFE2F3}0.42/-0.15         & \cellcolor[HTML]{CFE2F3}0.41/-0.19         & \cellcolor[HTML]{CFE2F3}0.41/-0.18         & \multicolumn{1}{c|}{\cellcolor[HTML]{CFE2F3}0.40/-0.18}         & \cellcolor[HTML]{FCE5CD}0.88/1.37          & \cellcolor[HTML]{FCE5CD}0.91/1.30          & \cellcolor[HTML]{FCE5CD}0.91/1.40          & \cellcolor[HTML]{FCE5CD}0.94/1.44          \\ \hline
\multicolumn{1}{c|}{DialoGPT-ft (l2r)}        & \cellcolor[HTML]{CFE2F3}0.43/-0.09         & \cellcolor[HTML]{CFE2F3}0.41/-0.04         & \cellcolor[HTML]{CFE2F3}0.47/0.01          & \multicolumn{1}{c|}{\cellcolor[HTML]{CFE2F3}0.46/0.00}          & \cellcolor[HTML]{FCE5CD}0.93/2.01          & \cellcolor[HTML]{FCE5CD}0.96/2.60          & \cellcolor[HTML]{FCE5CD}0.93/2.22          & \cellcolor[HTML]{FCE5CD}0.95/2.70          \\
\multicolumn{1}{c|}{BART-base-ft (s2s)}       & \cellcolor[HTML]{CFE2F3}0.37/-0.16         & \cellcolor[HTML]{CFE2F3}0.36/-0.14         & \cellcolor[HTML]{CFE2F3}0.37/-0.19         & \multicolumn{1}{c|}{\cellcolor[HTML]{CFE2F3}0.37/-0.13}         & \cellcolor[HTML]{FCE5CD}0.63/0.37          & \cellcolor[HTML]{FCE5CD}0.77/0.62          & \cellcolor[HTML]{FCE5CD}0.60/0.26          & \cellcolor[HTML]{FCE5CD}0.58/0.27          \\
\multicolumn{1}{c|}{BART-large-ft (s2s)}      & \cellcolor[HTML]{CFE2F3}0.36/-0.28         & \cellcolor[HTML]{CFE2F3}0.41/-0.13         & \cellcolor[HTML]{CFE2F3}0.35/-0.30         & \multicolumn{1}{c|}{\cellcolor[HTML]{CFE2F3}0.37/-0.23}         & \cellcolor[HTML]{FCE5CD}0.45/0.02          & \cellcolor[HTML]{FCE5CD}0.83/1.04          & \cellcolor[HTML]{FCE5CD}0.54/0.30          & \cellcolor[HTML]{FCE5CD}0.63/0.41          \\ \hline
\multicolumn{1}{c|}{Human}                    & \cellcolor[HTML]{CFE2F3}1.0                & \cellcolor[HTML]{CFE2F3}1.0                & \cellcolor[HTML]{CFE2F3}0.9                & \multicolumn{1}{c|}{\cellcolor[HTML]{CFE2F3}1.0}                & \cellcolor[HTML]{FCE5CD}1.0                & \cellcolor[HTML]{FCE5CD}1.0                & \cellcolor[HTML]{FCE5CD}1.0                & \cellcolor[HTML]{FCE5CD}1.0                \\ \hline
\end{tabular}
}
\caption{\textbf{Attribution probing results} for different response generation models on 4 dialogues datasets against two categories of corruptions.
}
\label{tab:att_results}
\end{table*}

We present results and findings for our two probing settings using different dialogue RG models across four datasets for which we collected verified explanations. For each human-validated explanation, we generate a corrupted version using one of our six corruption types, and compare the NLL for the probe target according to our two settings. 

In Tables~\ref{tab:inf_results} and~\ref{tab:att_results}, we show both binary accuracy and average difference in NLL for dialogues from four datasets under the two settings and aggregate the six corruption types into two categories. We also sample 5\% of the dialogues for human verification under the same two probe settings.

\subsection{How Does Probability of Response Change Given Explanations?}\label{Sec:inference_results}


\paragraph{All models are insensitive to the relation between explanations and responses.}
As shown in the left portion of Table~\ref{tab:inf_results}, we find that when comparing a valid explanation with a \emph{logically corrupted} (LC) one, all models, regardless of left-to-right or seq2seq model architecture, have accuracy around 50-60\%, near a random guessing baseline, with extremely small differences in NLL (some even negative). This suggests that the RG models do not understand the causal relation between the explanation and the response since they give similar probabilities to the response when conditioned on a valid explanation and on a incoherent explanation, while humans can easily identify the valid explanation.


\paragraph{Even gibberish does not change response probability much.}
Surprisingly, we find that even when corrupting the explanation so completely that it becomes unnatural English, most seq2seq RG models still generate responses with a roughly equal likelihood (left2right models perform better but still lag human performance) as shown in the right portion of Table~\ref{tab:inf_results}. \citet{sankar2019neural} find that the increase in perplexity of the response is tiny when they perturb the dialogue context, but here we find that there might even not be any increase in perplexity when conditioned on gibberish compared to a valid explanation expressed in English, while humans can identify the natural explanation perfectly. 

\subsection{Can RG Models Attribute Valid Reasons for the Responses?} \label{Sec:attribution_results}

\paragraph{Logically incoherent attribution confuses the models.}
Similar to the inference probing setting, for \emph{logically corrupted} one, all models have accuracy around 50-60\% and tiny differences in NLL from the left part of Table~\ref{tab:att_results}. This indicates that the RG models cannot identify a logically-valid reason for a response from a reason that is similarly natural in terms of grammar but with totally different and invalid logical implications for the dialogue. Humans, from our sampled dialogues, again show much higher accuracy in this setting. 

\paragraph{Models can confidently distinguish valid attribution from unnatural ones.}
For complete corruptions (CC), we find that except for BART, RG models perform much better in identifying a valid explanation compared to a completely corrupted one with most accuracy being close to 1 and relatively larger NLL differences. We conclude that these RG models find generating a valid explanation more natural than completely corrupted ones, which is expected since they are trained to generate natural sentences.
However, combining this finding with the previous observation, we find that these RG models can discern unnatural sentences by giving a low probability, but fail to determine the logical validity of the reasons for responses, posing doubts on whether they understand CSR behind a response.

\begin{figure}[t]
 	\centering
	\includegraphics[width=0.8\columnwidth]{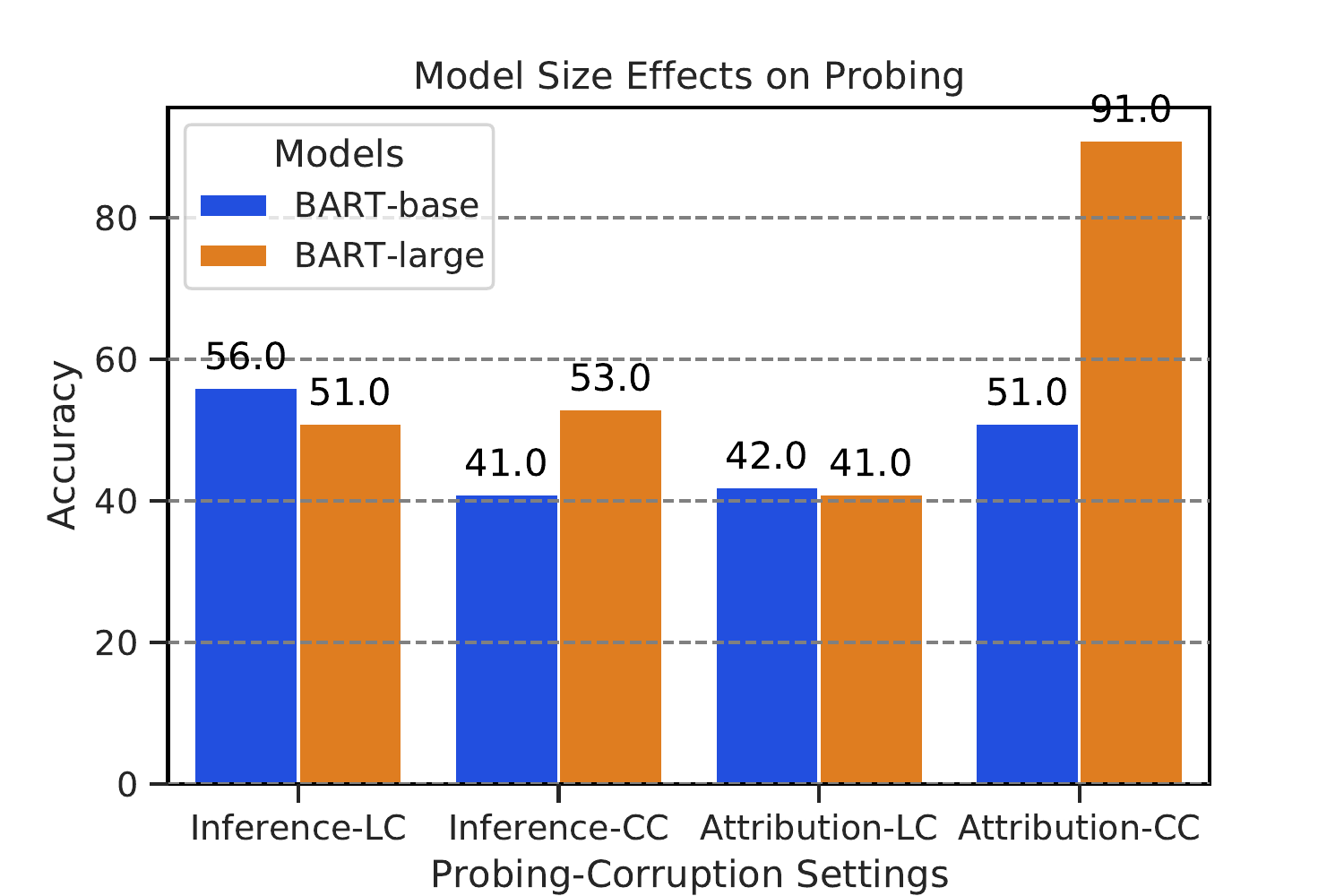}
	\caption{
	{\textbf{Model size Effects} on the two probing settings for BART aggregated across four datasets and types of corruptions. We find that except for the attribution setting against complete corruptions, increasing size does not impact much on probing performance.}}
	\label{fig:size_effects}
\end{figure}

\begin{figure}[t]
	\centering
	\includegraphics[width=\columnwidth]{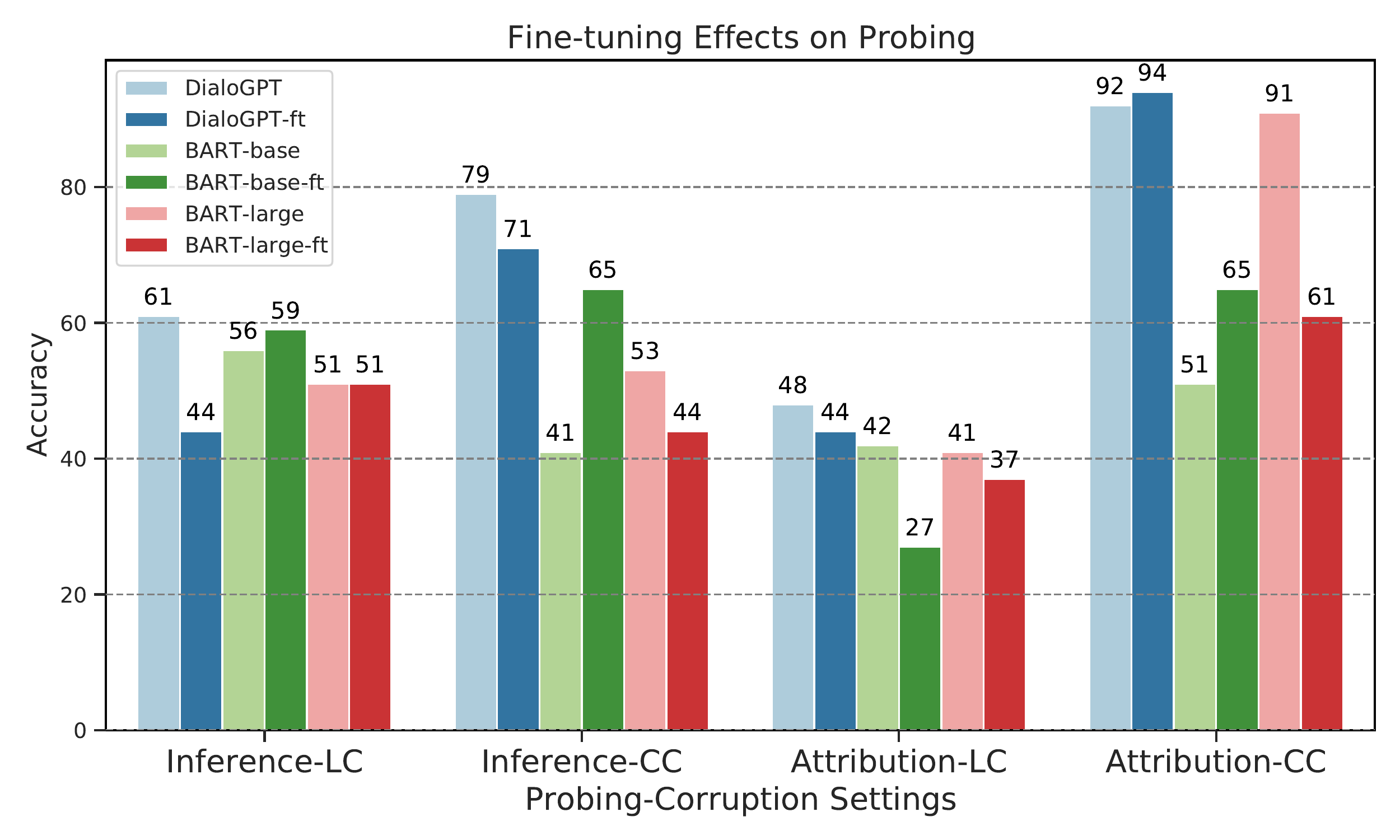}
	\caption{
	{\textbf{Fine-tuning (on in-domain dialogues) Effects} on the two probing settings for three models aggregated across four datasets and types of corruptions. We find that in general fine-tuning does not help and sometimes even hurt the probing performance.}}
	\label{fig:ft_effects}
\end{figure}

\subsection{Analysis of Probing Results} \label{Sec:analysis_results}

\paragraph{Unfamiliarity with probing format is not the bottleneck.}
Since these RG models are \emph{not} trained directly to take additional knowledge as input to generate responses or generate explanations for responses (although explaining happens often in dialogues), these poor results may be due to the probing setup. We thus fine-tune BART-base on 50\% of our verified explanations in the same format as our two settings and probe on the rest. We find even when the model is accustomed to the tasks, the accuracy against logical corruptions for both settings is still around 60\%. Although it is possible that with more data the performance can be improved, we also note that training with explanations also makes the model biased to prefer explanations over corrupted ones due to pattern matching. For example, no explanations contain negated connectives, which might be used to gain an advantage unrelated to understanding common sense when compared against negated corruption.

To probe a model that is accustomed to the task but not exposed to explanation patterns, we consider a GPT-2-based~\cite{radford2019language} model trained on TopicalChat~\cite{gopalakrishnan2019topical}, a knowledge-grounded dialogue dataset. The model is trained on given input of dialogue history concatenated with a knowledge sentence that the response needs to use. We treat the commonsense explanation as the knowledge sentence as they both provide necessary information that leads to the response. We find that the model performs similarly to DialoGPT on our probing setting for logical corruptions, providing evidence that the reason why these RG models cannot identify causal relations behind dialogue responses is not because the model is not used to taking explanation as input.

\paragraph{Model size does not help with understanding common sense.}
Comparing BART-base and BART-large in Figure~\ref{fig:size_effects}, we find that except for the attribution setting with complete corruptions, size does not change probing results (even lower accuracy against logical corruptions), indicating that the size of RG model is not the key to understand commonsense explanations for dialogue responses.

\begin{figure}[t]
    \hspace{-0.3cm}
	\includegraphics[width=\columnwidth]{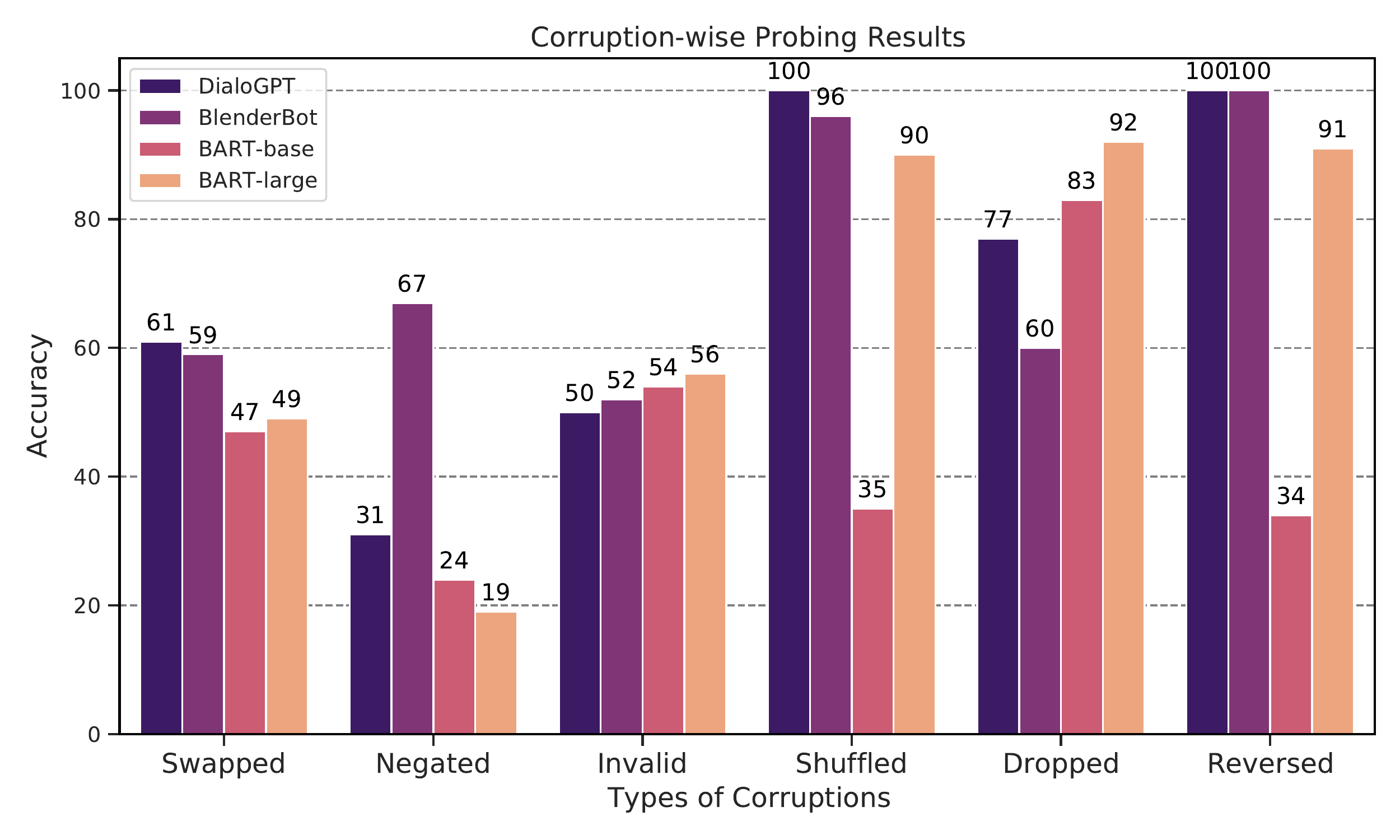}
	\caption{
	{\textbf{Corruption results} breakdown on the attribution probing settings aggregated across four datasets.}}
	\label{fig:corruption_wise}
\end{figure}

\paragraph{Fine-tuning on in-domain dialogues sometimes have opposite effects.}
Since these RG models are trained on different dialogue datasets that are not necessarily in the same domain as probing dialogues, we also explore the effects of fine-tuning on \emph{in-domain} dialogues, dialogues from the 4 datasets we use for probing. Three pairs of model (before and after fine-tuning) results are shown in Figure~\ref{fig:ft_effects} and we do not find significant differences. We even find sometimes fine-tuning hurts the probing results, which might be due to models picking up statistical patterns while training on similar dialogues, relying less on ``\emph{reasoning}'', if any.

\paragraph{Potential biases on certain perturbation types.}
The observations above are general trends of the models performance, but we also find interesting corner cases indicating potential biases in the models when we breakdown performance for six corruption types shown in Figure~\ref{fig:corruption_wise} on the attribution probing setting. For the \emph{Negated} corruption type, DialoGPT and BART have accuracy around 30\%, meaning that for 70\% of the time, they \emph{prefer} generating explanations with negated relations in it. 


\section{Related Work}\label{rel_work}
\paragraph{Commonsense Reasoning}
The majority of recent CSR benchmarks~\cite{zellers2018swagaf, talmor2019commonsenseqa,bisk2019piqa,sap2019social, lin-etal-2021-riddlesense, lin-etal-2021-xcsr, lin2020commongen}  test a model's ability to choose the correct option given a context and a question. 
Recent work also aims to probe models in these tasks to see if reasoning is actually achieved~\cite{richardson2020does,richardson2020probing,zhou2020rica, lin-etal-2021-differentiable}.
~\citet{arabshahi2020conversational} focuses on if-then-because reasoning in conversations and design a theorem prover.
In RG, several works have tried to incorporate commonsense~\cite{zhou2018commonsense, zhang2019grounded} using ConceptNet, a commonsense knowledge graph~\cite{liu2004conceptnet} to make responses more natural-sounding.

\paragraph{Dialogue Response Generation}
Recent work focused on fine-tuning large pre-trained transformer models~\cite{radford2019language,zhang2020dialogpt} on dialogue data. 
Many dialogue datasets have been collected with different focuses such as incorporating knowledge~\citep{gopalakrishnan2019topical,dinan2018wizard}, empathy~\cite{rashkin2019towards}, personality~\cite{zhang2018personalizing} and reasoning~\cite{cui2020mutual} within dialog systems. 
There has also been work on combining a variety of datasets to exhibit multiple attributes~\cite{roller2020recipes}.

\section{Conclusion}\label{conclusion}
We study commonsense reasoning in dialogue response generation aiming to close the gap between current RG models and human communication abilities. Specifically we formalize the problem by framing commonsense as a latent variable in the RG task and using explanations for responses as textual form of commonsense. We design an explanation collection procedure for RG and propose two probing settings to evaluate RG models' CSR capabilities. We hope our study motivates more research in making RG models emulate human reasoning process in pursuit of smooth human-AI communication.

\section*{Acknowledgments}
We thank anonymous reviewers for providing insightful feedback along with Brendan Kennedy, Peifeng Wang, and members from INK and JAUNTS lab. This research is supported in part by the DARPA MCS program under Contract No. N660011924033, the Defense Advanced Research Projects Agency with award W911NF-19-20271, NSF IIS 2048211, and NSF SMA 182926.

\section*{Ethics and Broader Impact}
Our work aims to examine RG model's ability to understand common sense for dialogue responses. ~\citet{sheng2021nice} have found biases in DialoGPT responses and ~\citet{mehrabi2021lawyers} have found representational harms in common sense resources. We acknowledge that the generated responses from models we use in probing experiments might contain biases.
All of the dialogue datasets and models are in English, which benefits English speakers more. We have conducted human verification using Amazon Mechanical Turks. We pay turkers around \$14 per hour, well above the highest state minimum wage and engage in constructive discussions if they have concerns about the process. We also give each annotation instance enough time so that we do not pressure annotators.


\bibliography{anthology,custom_rebiber}
\bibliographystyle{acl_natbib}

\clearpage

\appendix
\section{GLUCOSE Detail}\label{glucose}
\begin{figure}[]
	\includegraphics[width=\columnwidth]{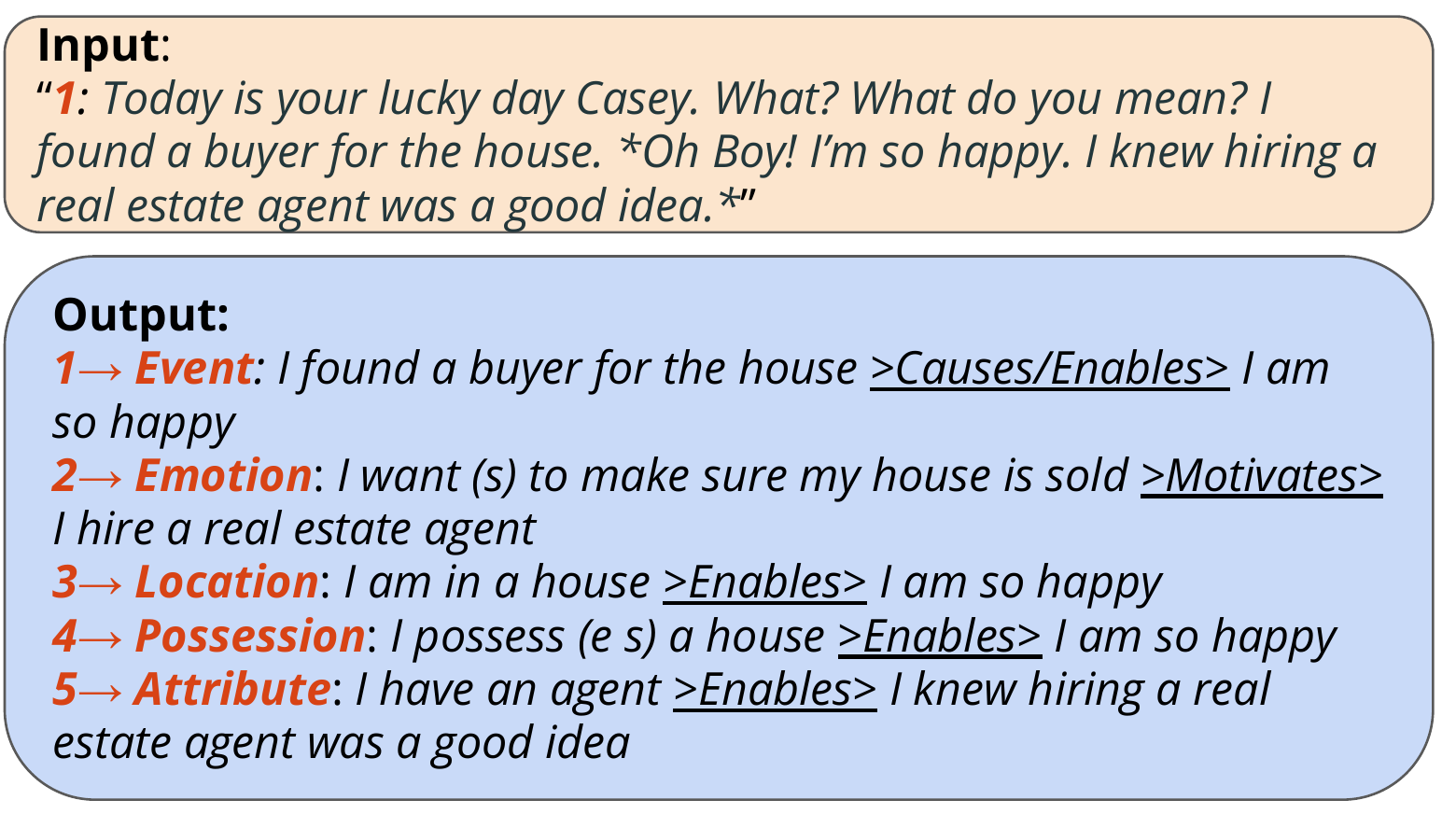}
	\caption{
	{\textbf{Example input and output} from GLUCOSE-trained T5 model on a dialogue.}}
	\label{fig:GLUCOSE}
\end{figure}

GLUCOSE contains human annotations of ten dimensions of causal explanation related to X. 
Five of the dimensions are about events and states happening before X and five are about those happening after X. 
Specifically, inspired by cognitive psychology, the authors of GLUCOSE consider events, emotions, location states, possession states, and other attributes as the five dimensions of causal inferences.
According to their evaluation, the best-performing model is T5~\cite{raffel2020exploring} (with 770M parameters) with the input formulated as $\#d: S^\ast[X]$, where $d$ is the dimension and $S^\ast[X]$ is the story $S$ with sentence $X$ surrounded by asterisks''. An illustrated example of inputs of outputs of the T5 model trained on GLUCOSE is shown in Figure~\ref{fig:GLUCOSE}.

To adopt the T5 model trained on GLUCOSE to our task: generating explanations about what might cause producing a response given a dialogue history, we append the dialogue history turns together, enclose the response we are interested in explaining with asterisks, and fill in dimension number 1 to 5 to ask for what event, emotion, location, possession, and attribute could cause, motivate, or enable the response. In other words, we formulate our queries as $\#d: H^\ast[R]$, where $d$ is the dimension 1 to 5 and $H^\ast[R]$ is the dialogue history $H$ appended with the response $R$ surrounded with asterisks.

\section{Verification Detail}\label{sec:veri_detail}
\begin{figure}[t]
	\includegraphics[width=0.8\columnwidth]{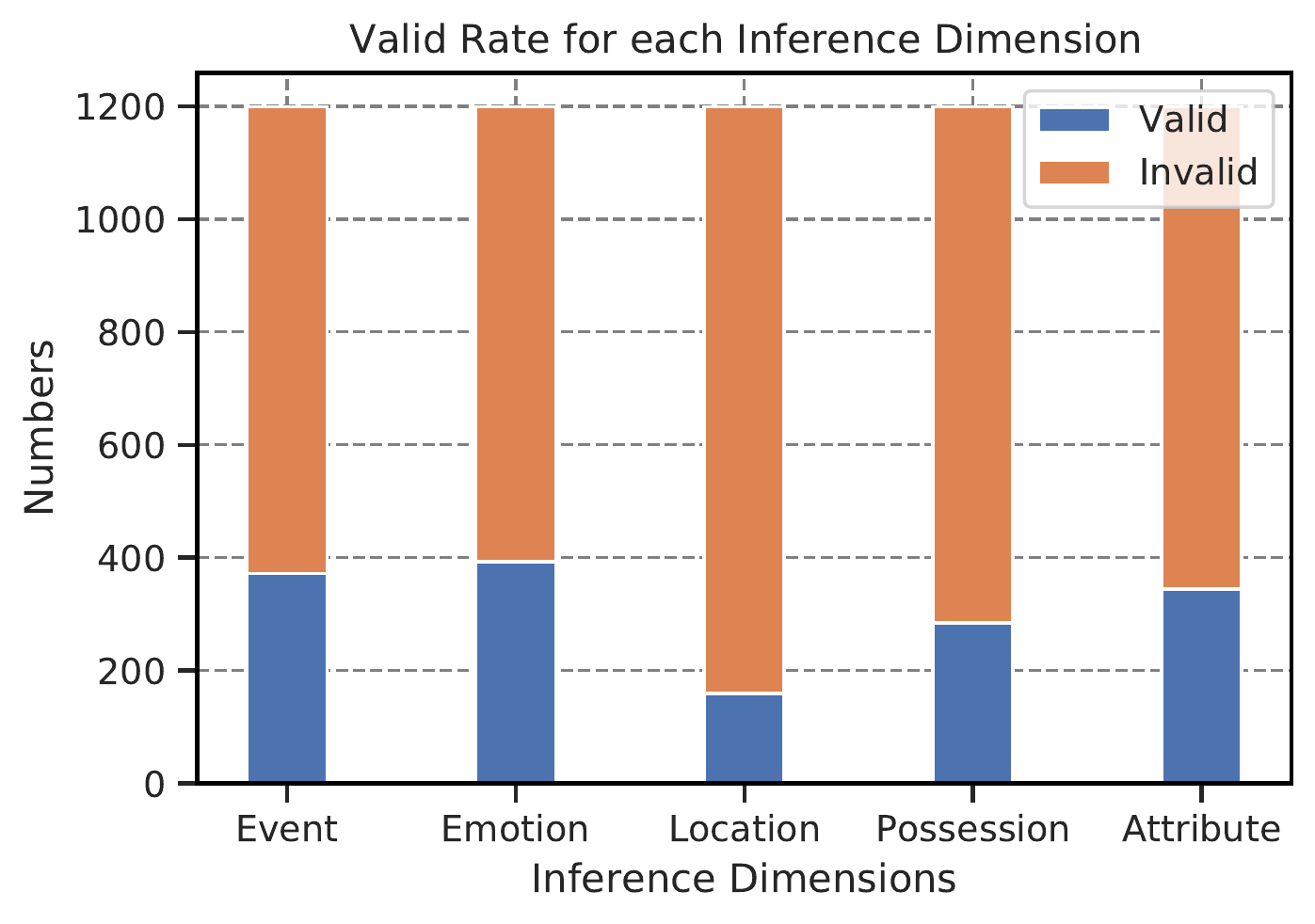}
	\caption{
	{\textbf{Verification results} on 6k explanations from 1,200 dialogeus separated by five causal dimensions. The valid rates are 31\%, 33\%, 13\%, 24\%, and 29\% for the five dimensions.}
	}
	\label{fig:veri_results}
\end{figure}

Table~\ref{tab:pass_rate} shows the general pass rate for each criterion and the overall pass rate (need to pass all three criteria).
Figure~\ref{fig:veri_results} shows distribution of valid and invalid explanations separated by the five causal dimensions. We find the explanations about a \emph{location} state that causes the response have a lower valid rate (13\%) than others. This might be due to that in some dialogues the location information is not important in explaining the response and thus it is difficult to come up with a plausible reason about a location that leads to the response. All other dimensions have a similar rate of 25-30\%.

To ensure annotation quality, the workers first need to pass a \emph{qualification test} (QT) that tests their understandings of the criteria to be able to do our main annotation tasks. Our QT contains eight questions, each contains a dialogue history, a response, and an explanation and we ask them to choose whether this explanation satisfies a specific criterion from the three above. The eight questions are formed into 4 pairs each consisting of a \emph{training} question and a \emph{testing} question and each pair focuses on the same criterion. For the \emph{relevance} and \emph{non-trivial} criteria, we have one pair for each and for the \emph{plausible} criterion, we have two pairs since it is trickier to determine than the other two. We provide the right answer with explanation for the training question whether they answer it correctly or not and use the testing questions for assessment of their understanding.
\begin{table}[]
\centering
\resizebox{0.5\columnwidth}{!}{
\begin{tabular}{c|c}
\hline
\textit{\textbf{Criterion}} & \textit{\textbf{Passing Rate}} \\ \hline
Relevant                    & 55\%                           \\ \hline
Non-trivial                 & 73\%                           \\ \hline
Plausible                   & 37\%                           \\ \hline
All three                   & 26\%                           \\ \hline
\end{tabular}
}
\caption{Passing rates for the three criteria and the overall valid rate (need to pass all three) from verification. }
\label{tab:pass_rate}
\vspace{-0.5cm}
\end{table}
\begin{figure*}[h]
\centering
\includegraphics[width=\textwidth]{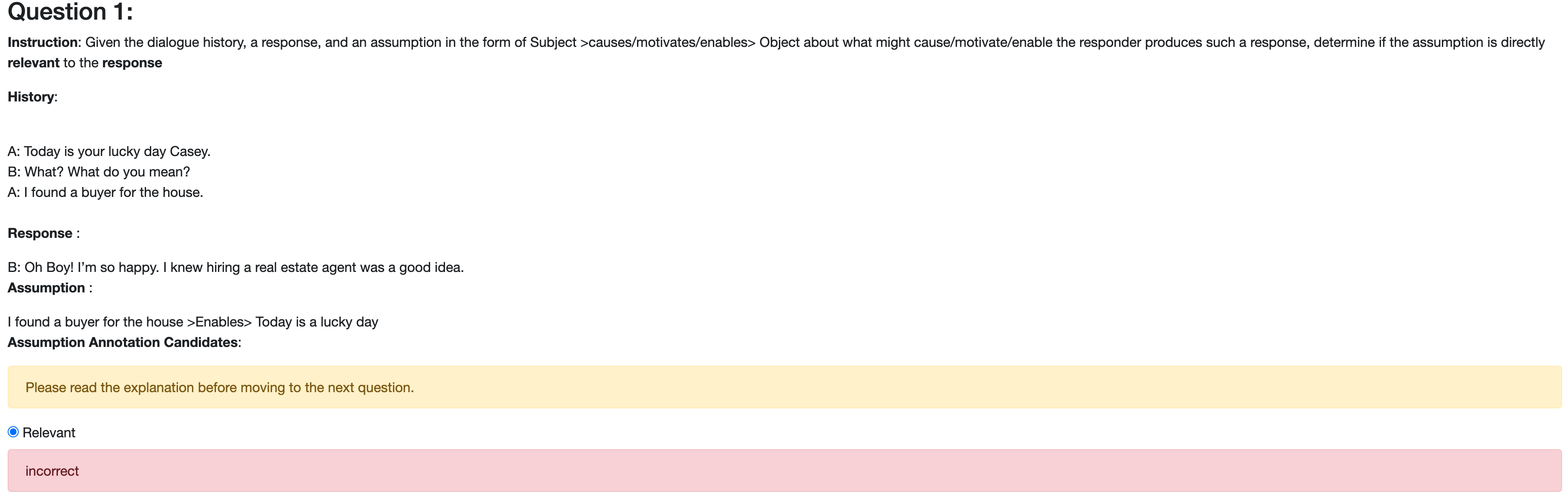}
\caption{Example of qualification test question with shown explanation.}
\label{fig:QT_UI}
\end{figure*}

\begin{figure*}[h]
\centering
\includegraphics[width=\textwidth]{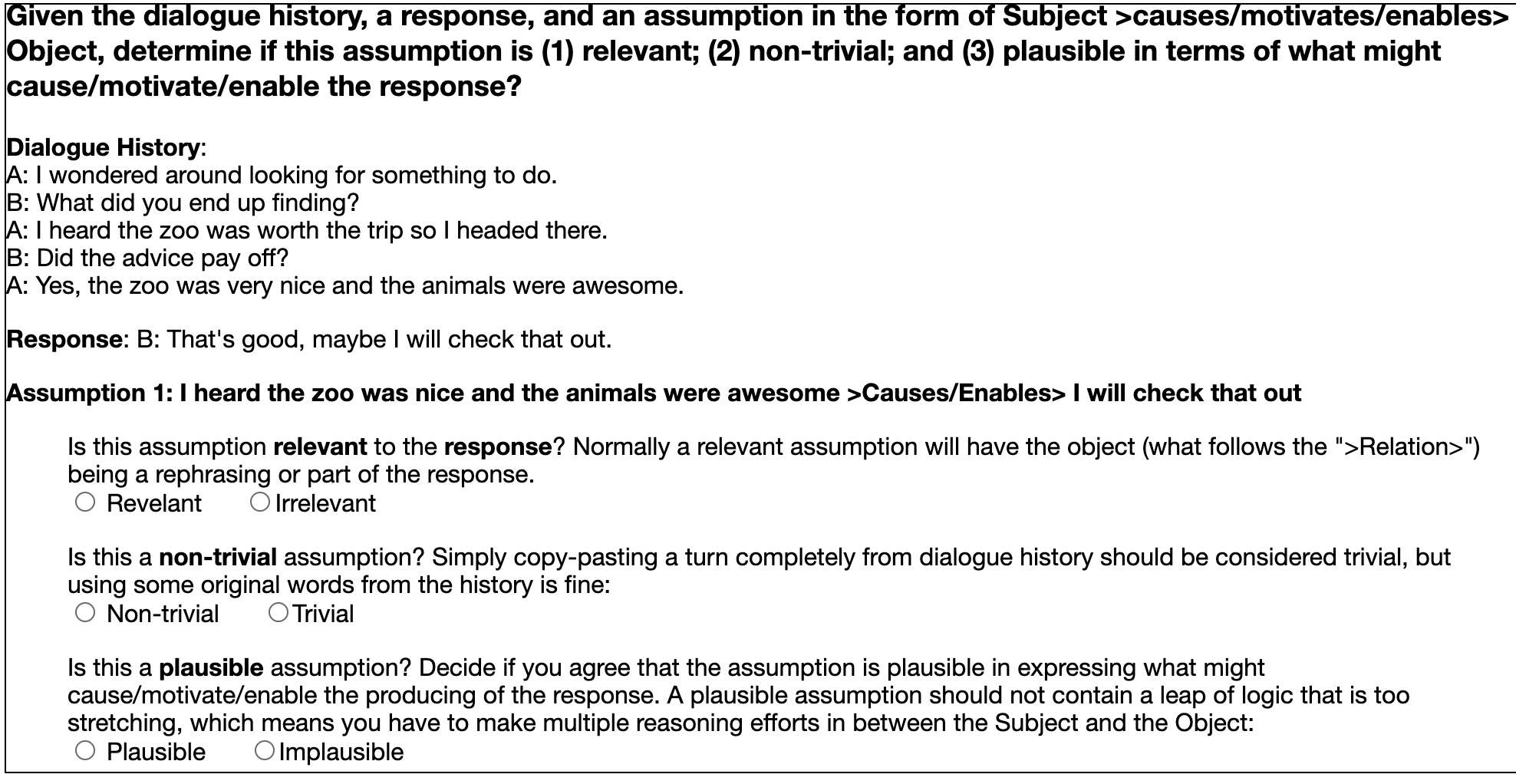}
\caption{Example of the verification task question with three criteria for verifiers to choose.}
\label{fig:MT_UI}
\end{figure*}

\begin{figure}[t]
	\includegraphics[width=\columnwidth]{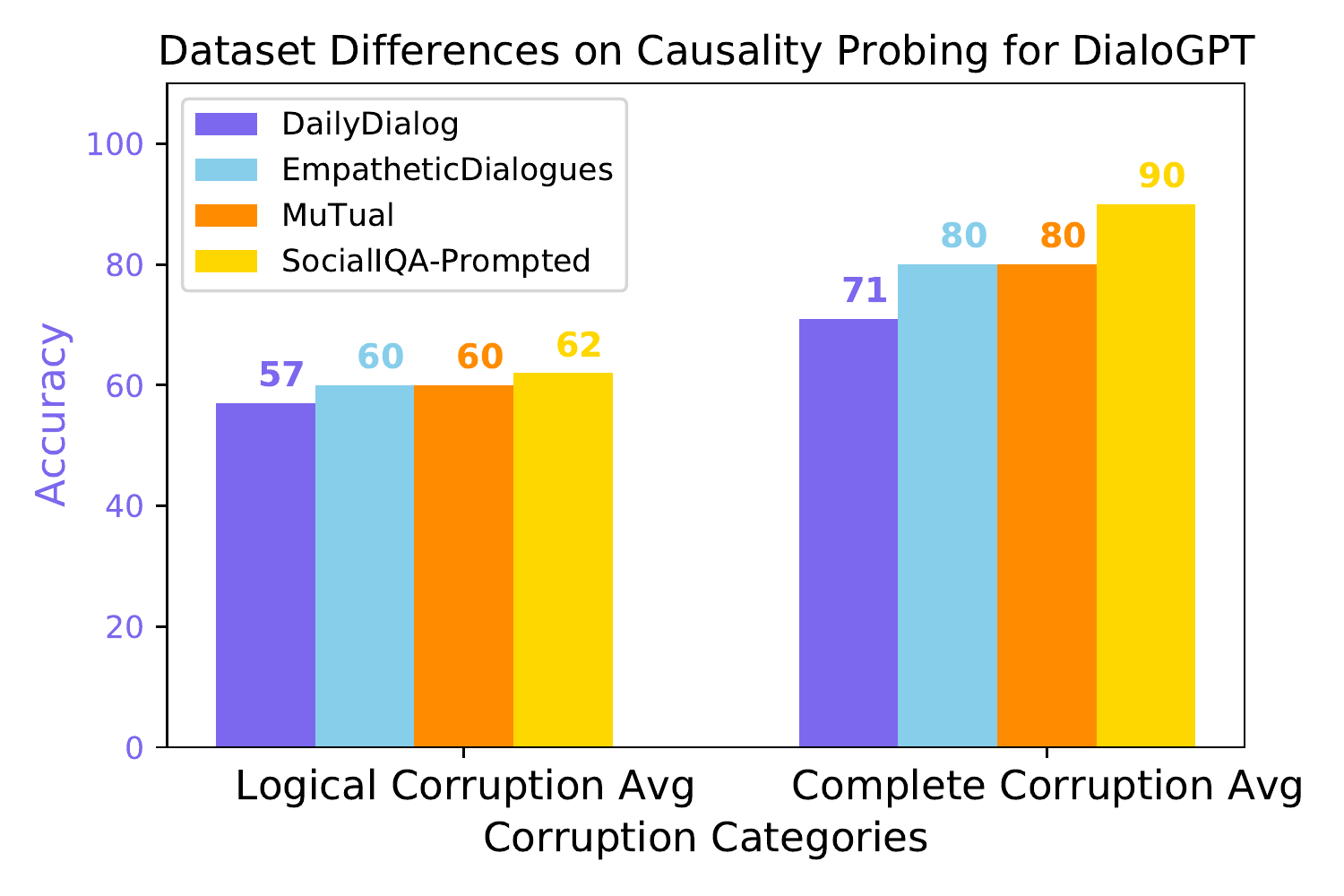}
	\caption{
	{\textbf{Dataset differences} on the causality probing setting for DialoGPT model, we find that there is no drastic differences in probing performances across four datasets for logical corruptions, i.e., the conclusion that RG model fails to understand causality holds true for all datasets. We see difference in accuracy ranging from 70\% to 90\% for complete corruptions.}}
	\label{fig:dataset}
\end{figure}

\section{Model Detail}\label{model_detail}

\paragraph{DialoGPT} extends the GPT-2 architecture that adopts the generic transformer language model~\cite{vaswani2017attention} by training on 147M multi-turn conversation-like exchanges extracted from Reddit. We use the 345M DialoGPT model\footnote{\scriptsize \url{https://huggingface.co/microsoft/DialoGPT-medium}}~\cite{zhang2020dialogpt}.

\paragraph{BlenderBot} is proposed by ~\citet{roller2020recipes} using a standard seq2seq transformer architecture~\cite{vaswani2017attention}. The model aims to blend multiple conversational skills.  Human evaluations show their best models beat existing approaches in multi-turn dialogue in terms of engagingness and humanness. We use the 400M BlenderBot model distilled from 2.7B parameter model\footnote{\scriptsize \url{https://huggingface.co/facebook/blenderbot-400M-distill}}.

\paragraph{BART} is proposed by~\citet{lewis2019bart} using a standard seq2seq architecture with a bidirectional BERT encoder and a left-to-right GPT decoder. It uses denoising pre-training objectives and has shown to outperform previous models in multiple language generation tasks including ConvAI2~\cite{dinan2019second}. We use both BART-base and BART-large with 139M and 406M parameters, respectively\footnote{\scriptsize \url{https://huggingface.co/models?search=bart}}.

\end{document}